\documentclass{article}
\PassOptionsToPackage{numbers, compress}{natbib}

\usepackage[final]{neurips_2024_ml4ps}

\usepackage[utf8]{inputenc} 
\usepackage[T1]{fontenc}    
\usepackage{hyperref}       
\usepackage{url}            
\usepackage{booktabs}       
\usepackage{amsfonts}       
\usepackage{nicefrac}       
\usepackage{microtype}      
\usepackage{xcolor}         
\usepackage{amsmath}
\usepackage{graphicx}
\usepackage{subcaption}
\title{Data-Driven, Parameterized Reduced-order Models for Predicting Distortion in Metal 3D Printing}

\author{%
   Indu Kant Deo \\
  Department of Mechanical Engineering\\ 
  The University of British Columbia\\
  Vancouver, BC V6T 1Z4\\
  \texttt{indukant@mail.ubc.ca} \\
  \And
  Youngsoo Choi \\ 
  Lawrence Livermore National Laboratory \\
  Livermore, CA, 94550 \\
  \texttt{choi15@llnl.gov} \\
  \And
  Saad A. Khairallah \\ 
  Lawrence Livermore National Laboratory \\
  Livermore, CA, 94550 \\
  \texttt{khairallah1@llnl.gov} \\
  \And
    Alexandre Reikher \\ 
  Lawrence Livermore National Laboratory \\
  Livermore, CA, 94550 \\
  \texttt{reikher1@llnl.gov} \\
  \And
    Maria Strantza \\ 
  Lawrence Livermore National Laboratory \\
  Livermore, CA, 94550 \\
  \texttt{strantza1@llnl.gov} \\
}

\begin{document}

\maketitle

\begin{abstract}

In Laser Powder Bed Fusion (LPBF), the applied laser energy produces high thermal gradients that lead to unacceptable final part distortion. Accurate distortion prediction is essential for optimizing the 3D printing process and manufacturing a part that meets geometric accuracy requirements. This study introduces data-driven parameterized reduced-order models (ROMs) to predict distortion in LPBF across various machine process settings. We propose a ROM framework that combines Proper Orthogonal Decomposition (POD) with Gaussian Process Regression (GPR) and compare its performance against a deep-learning based parameterized graph convolutional autoencoder (GCA). The POD-GPR model demonstrates high accuracy, predicting distortions within $\pm0.001mm$, and delivers a computational speed-up of approximately 1800x.
\end{abstract}

\section{\label{sec:introduction}Introduction}

LPBF is a popular metal additive manufacturing technique that has gained significant attention in recent years due to its ability to fabricate complex geometries with high precision. 
In LPBF, a thin layer of metal powder is deposited on a build platform, and a high-energy laser selectively melts and fuses the powder particles together to form a solid layer. 
This process is repeated layer by layer, with each 2D layer fusing to the previous one, ultimately constructing fully dense 3D components \cite{gibson2015}.

The repeated melting and solidification cycles in LPBF lead to significant thermal gradients, resulting in notable distortion in the as-built part. 
This distortion can compromise the dimensional accuracy and structural integrity of the final component, which is a critical requirement in many applications \cite{debroy2018}. 
To address this challenge, currently, the additive manufacturing community relies on a trial-and-error method, which involves conducting numerous experiments or simulations that are time-consuming and expensive. The approach is a distortion compensation technique, which involves pre-distorting the part design in such a way that upon printing, the final built shape matches the intended geometry \cite{vayre2012}. The problem is predicting quickly the pre-distorted part geometry is not easy.

The amount of distortion for a given geometry depends on various machine settings, such as scan speed, laser power, and dwell time \cite{king2015}. 
Accurately predicting the distortion based on these parameters is crucial for effective distortion compensation and process optimization \cite{paudel2023physics}.
High-fidelity finite element models offer a cost-effective alternative, enabling repeated trials without the need to physically build the parts \cite{dunbar2016experimental}.
However, developing accurate physics-based models for LPBF distortion is a complex task due to the non-linear dependence of distortion on the process and complex part geometries: one pre-distortion solution that fits all problems does not exist \cite{yadollahi2017}.

This study aims to develop parameterized data-driven reduced-order models (ROMs) for accurately predicting distortion in the LPBF process under various machine settings \cite{fries2022lasdi,dong2024data, brown2023data}. 
Specifically, it employs a combination of POD and Gaussian Process Regression to create a ROM, which is then compared with a parameterized graph convolutional autoencoder for distortion prediction. 
The POD-GPR ROM achieves a distortion prediction accuracy within $\pm 0.001 , \text{mm}$, and offers a computational speed-up over the high fidelity model of nearly $1800$ times. 
This significant improvement highlights the model's potential for enabling rapid and precise distortion predictions, which is critical for optimizing LPBF processes. 
The ability to efficiently predict distortion not only reduces the reliance on costly and time-consuming experimental trials but also enhances the overall process control, making this approach highly valuable for industrial applications \cite{parry2016}.

\section{\label{sec:methods}Methods}

\subsection{LPBF simulation data}
\begin{figure}[ht]
    \centering
    \includegraphics[width=0.99\linewidth]{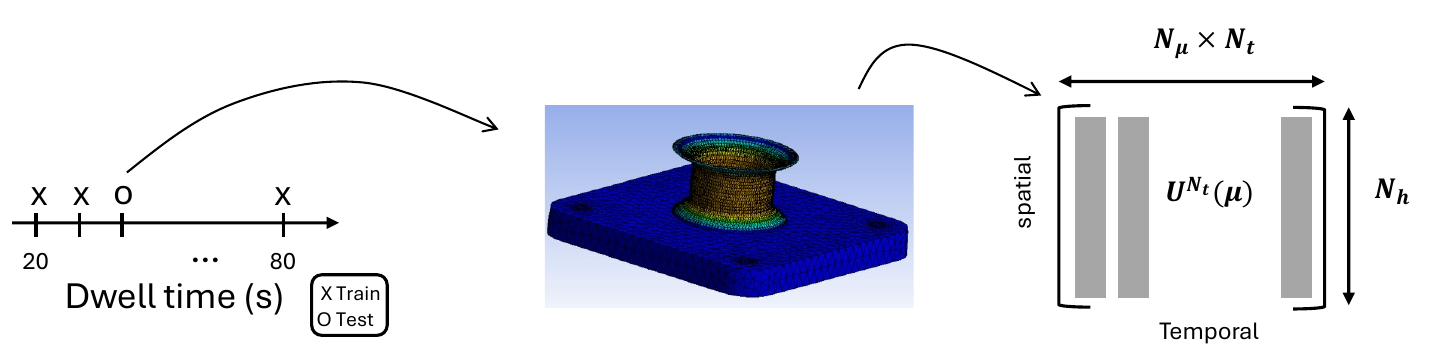}
    \caption{Schematic of dataset generation. LPBF simulations were generated for parameter dwell time. Distortion data was extracted for each simulation and arranged into a training snapshot matrix.}
\label{fig:fig1}
\end{figure}

In this study, we analyze data generated from Laser Powder Bed Fusion (LPBF) simulations conducted using ANSYS$^\circledR$ Additive Suite. 
The dataset is parameterized based on the dwell time $(dt)$, which represents the time interval required for the laser or heat source to revisit a specific location to deposit a subsequent material layer \cite{foster2017impact}. 
The impact of this interlayer dwell time becomes more pronounced as structures grow in size and complexity, significantly affecting the thermal history and geometric distortion outcomes. 
Specifically, we performed simulations on a cylindrical geometry with dwell times $dt \in [20, 80] \, \mathrm{s}$, sampled at intervals of $5 \, \mathrm{s}$. 
The computational mesh used in these simulations comprised $N_h = 77,151$ nodes, and each simulation covered 34 layers of metal deposition, yielding $N_t = 34$ time steps per simulation. 
On average, each simulation required approximately 2 hours of computation on 112 cores of an Intel(R) Xeon(R) CLX-8276L processor. 
From this extensive dataset, we selected $N_\mu$ samples for training, with the remaining samples designated for validation and testing. 
Let $\boldsymbol{\mu}^{(i)} = \left( dt^{(i)} \right)$ represent the $i$-th parameter in the training set. 
The primary quantity of interest is the final distortion field, $\mathbf{u}$. 
We denote the snapshot vector of distortion data at time-step $n$ for parameter $\boldsymbol{\mu}^{(i)}$ as $\mathbf{u}_n^{(i)} \in \mathbb{R}^{N_h}$, and the corresponding data matrix for all nodes and time steps as $\mathbf{U}^{(i)} = \left[\mathbf{u}_0^{(i)}, \dots, \mathbf{u}_{N_t}^{(i)}\right] \in \mathbb{R}^{N_h \times N_t}$. 
By aggregating all these matrices, we form a third-order tensor $\mathbf{U} = \left[\mathbf{U}^{(1)}, \dots, \mathbf{U}^{(N_\mu)}\right] \in \mathbb{R}^{N_\mu \times N_h \times N_t}$, which constitutes the training dataset utilized in this work. A visual representation of this dataset is provided in Figure \ref{fig:fig1}.

\subsection{POD-GPR}
POD-GPR is a parameterized, data-driven reduced order modeling method which consists of two main features: 1) POD \cite{berkooz1993proper} to learn a linear spatial compression of data to a latent space; 2) GPR \cite{schulz2018tutorial} to map the POD coefficients of final distortion layer to a given parameter value.

POD is a dimensionality reduction technique widely used in the analysis of complex systems to identify dominant patterns or modes. 
Given a set of snapshots $\mathbf{U} = [\mathbf{U}_1, \mathbf{U}_2, \dots, \mathbf{U}_{N_t}] - \mathbf{U}_{ref} \in \mathbb{R}^{N_h \times N_t}$, where each snapshot $\mathbf{U}_i \in \mathbb{R}^{N_h}$ represents the state of the system at a particular time or parameter setting, the goal of POD is to find a set of orthonormal basis vectors $\{\mathbf{\phi}_k\}_{k=1}^r$ that capture the most energetic features of the data. This is achieved by solving the eigenvalue problem associated with the covariance matrix $\mathbf{C} = \mathbf{U}\mathbf{U}^T \in \mathbb{R}^{{N_h} \times {N_h}}$. The eigenvectors corresponding to the largest eigenvalues provide the POD modes, which minimize the reconstruction error in a least-squares sense. Mathematically, this can be expressed as:

\[
\min_{\mathbf{\phi}_k} \sum_{i=1}^{N_t} \left\| \mathbf{U}_i - \sum_{k=1}^r \langle \mathbf{U}_i, \mathbf{\phi}_k \rangle \mathbf{\phi}_k \right\|^2,
\]

where $\langle \mathbf{U}_i, \mathbf{\phi}_k \rangle$ denotes the projection of $\mathbf{U}_i$ onto the $k$-th POD mode. The reduced representation of the data is then given by the projection coefficients $\mathbf{a}_i = [\langle \mathbf{U}_i, \mathbf{\phi}_1 \rangle, \dots, \langle \mathbf{U}_i, \mathbf{\phi}_r \rangle]^T$, enabling efficient analysis and computation in a lower-dimensional subspace. We select the first r modes for which the total energy is greater than $99.99\%$. The energy stored in the first r modes is given by:
\[
E_r = \frac{\sum_{j=1}^r \sigma_j^2}{\sum_{j=1}^N \sigma_j^2},
\]
where $\sigma$ is the eigenvalue. 

GPR is a powerful non-parametric method for modeling complex functions, particularly in cases where uncertainty quantification is important. 
In this work, we employ GPR to learn $r$ independent Gaussian process regressions, each mapping the final time POD coefficients to their corresponding parameter $\boldsymbol{\mu}$. Let $\mathbf{a}(\boldsymbol{\mu}) = [a_1(\boldsymbol{\mu}), \dots, a_r(\boldsymbol{\mu})]^T$ denote the vector of POD coefficients at the final time for the parameter $\boldsymbol{\mu}$. For each coefficient $a_j(\boldsymbol{\mu})$, where $j \in \{1, \dots, r\}$, we model it as a Gaussian process:

\[
a_j(\boldsymbol{\mu}) \sim \mathcal{GP}(m_j(\boldsymbol{\mu}), k_j(\boldsymbol{\mu}, \boldsymbol{\mu}')),
\]

where $m_j(\boldsymbol{\mu})$ is the mean function and $k_j(\boldsymbol{\mu}, \boldsymbol{\mu}')$ is the covariance function (kernel) associated with the $j$-th POD coefficient. In our approach, we use a constant mean function for \( m_j(\boldsymbol{\mu}) \) and a radial basis function (RBF) kernel for \( k_j(\boldsymbol{\mu}, \boldsymbol{\mu}') \), which is defined as:

\[
k_j(\boldsymbol{\mu}, \boldsymbol{\mu}') = \sigma_j^2 \exp\left(-\frac{\|\boldsymbol{\mu} - \boldsymbol{\mu}'\|^2}{2\ell_j^2}\right),
\]

where \( \sigma_j^2 \) is the signal variance, and \( \ell_j \) is the length-scale parameter. 

Given training data $\mathcal{D} = \{(\boldsymbol{\mu}^{(i)}, \mathbf{a}^{(i)})\}_{i=1}^{N_\mu}$, the posterior distribution for each $a_j(\boldsymbol{\mu}^*)$ at a new test point $\boldsymbol{\mu}^*$ can be computed as:

\[
a_j(\boldsymbol{\mu}^*) \mid \mathcal{D}, \boldsymbol{\mu}^* \sim \mathcal{N}(\hat{m}_j(\boldsymbol{\mu}^*), \hat{k}_j(\boldsymbol{\mu}^*, \boldsymbol{\mu}^*)),
\]

where $\hat{m}_j(\boldsymbol{\mu}^*)$ and $\hat{k}_j(\boldsymbol{\mu}^*, \boldsymbol{\mu}^*)$ are the posterior mean and variance, respectively. This approach allows us to efficiently predict the POD coefficients for unseen parameters $\boldsymbol{\mu}$, capturing the underlying uncertainty in the process.

\subsection{Parameterized graph-convolution autoencoder}
\begin{figure*}
    \centering
    \includegraphics[width=0.8\textwidth]{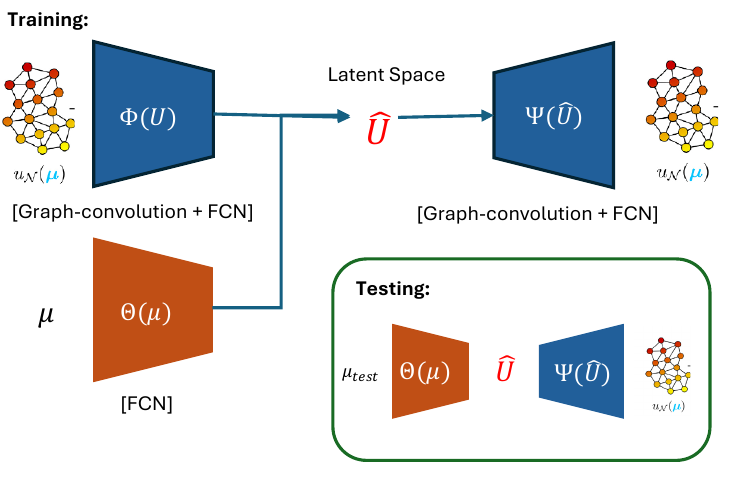}
    \caption{Illustration of parameterized graph convolutional autoencoder architecture }
\label{fig:Gca}
\end{figure*}

We employ a graph convolutional autoencoder (GCA) \cite{pichi2024graph, lee2024virtual, gao2024finite} to map a graph-based representation of the distortion field into a latent space. 
The encoder, $\mathcal{E}$, applies graph convolution layers to extract features and reduce dimensionality, resulting in a latent representation $\mathbf{\hat{U}} = \mathcal{E}(\mathbf{X}; \Theta_e)$, where $\mathbf{X}$ denotes the input graph and $\Theta_e$ the encoder parameters.
Simultaneously, we train a fully connected neural network (FCNN), $\mathcal{F}$, that maps the dwell time $dt$, to this same latent space, $\mathbf{\hat{U}}^{(p)} = \mathcal{F}(dt; \Theta_f)$, with $\Theta_f$ representing the network parameters. This setup ensures that the latent space encodes both the geometrical and operational characteristics influencing the distortion.
The decoder, $\mathcal{D}$, then reconstructs the distortion field from the latent space, $\mathbf{\hat{X}} = \mathcal{D}(\mathbf{\hat{U}}; \Theta_d)$, aiming to minimize the reconstruction loss $\mathcal{L}_{rec} = \|\mathbf{X} - \mathbf{\hat{X}}\|^2$. Figure \ref{fig:Gca} depicts the architecture of parameterized graph convolutional autoencoder.

The overall training objective also includes a parameter consistency loss $\mathcal{L}_{param} = \|\mathbf{\hat{U}} - \mathbf{\hat{U}}^{(p)}\|^2$, ensuring that the FCNN's output aligns with the encoder's latent space.
The mathematical formulation of our training objective combines these losses, optimizing $\Theta_e$, $\Theta_f$, and $\Theta_d$ through:

\[
\min_{\Theta_e, \Theta_f, \Theta_d} \mathcal{L}_{rec}(\mathbf{X}, \mathbf{\hat{X}}) + \lambda \mathcal{L}_{param}(\mathbf{\hat{U}}, \mathbf{\hat{U}}^{(p)}),
\]

where $\lambda$ is a regularization parameter balancing the two loss components.  For the present work, we have used $\lambda = 0.5$.
\section{\label{sec:results}Results}
In this study, the training dataset consisted of dwell times $dt = \{20, 25, 35, 40, 50, 55, 65, 70, 80\}$ seconds. The present study aims to test the performance of data-driven models in a scarce data regime. Validation was conducted using dwell times $dt = \{30, 60\}$ seconds, while testing was performed on $dt = \{45, 75\}$ seconds. In particular, for the POD-GPR model, there is no separate validation set; both validation and test sets are combined into a single test set. The current selection of validation and test sets focuses on evaluating the performance of these models within the interpolation range of the parameters. In future work, we plan to extend the parameter space to include the extrapolation range, enabling a more comprehensive assessment of the models' generalization capabilities.

\begin{figure}[ht]
    \centering
    \includegraphics[width=0.99\linewidth]{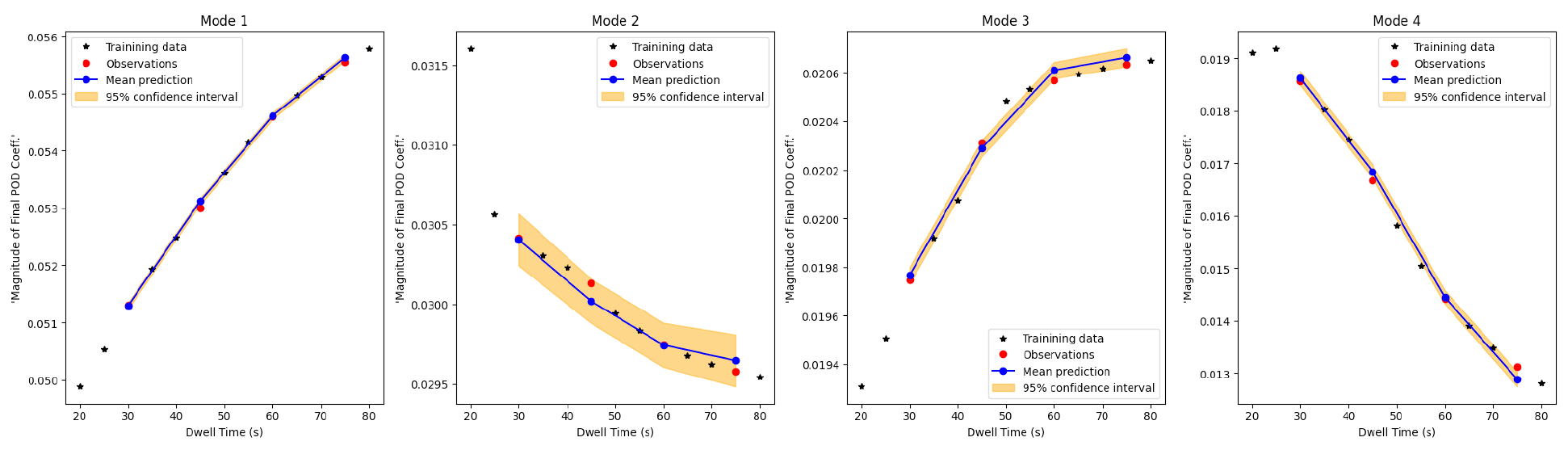}
    \caption{Graphical representation of the first four POD coefficients of final layer predicted by the GPR model for various dwell times in the test set ($dt = \{30, 45, 60, 75\}$ seconds) along with the 95\% confidence interval.}
\label{fig:fig2}
\end{figure}

Our implementation of POD successfully preserved 99.99\% of the variance with 129 modes. Subsequently, 129 independent GPRs were trained. The performance of the first four POD coefficients corresponding to the test dwell times is visually represented in the Figure \ref{fig:fig2}.

The industry standard for accuracy in additive manufacturing processes is within $\pm0.1$ mm. Our POD-GPR model significantly exceeds this requirement, achieving an accuracy of $\pm0.001$ mm for the maximum displacement value, showcasing an excellent agreement with the finite element simulations. Notably, the runtime for the POD-GPR is approximately 4 seconds, providing a computational speed-up of about 1800 times compared to traditional finite-element methods.

\begin{figure}[ht]
    \centering
    \begin{subfigure}{0.49\textwidth}
        \centering
        \includegraphics[width=\textwidth]{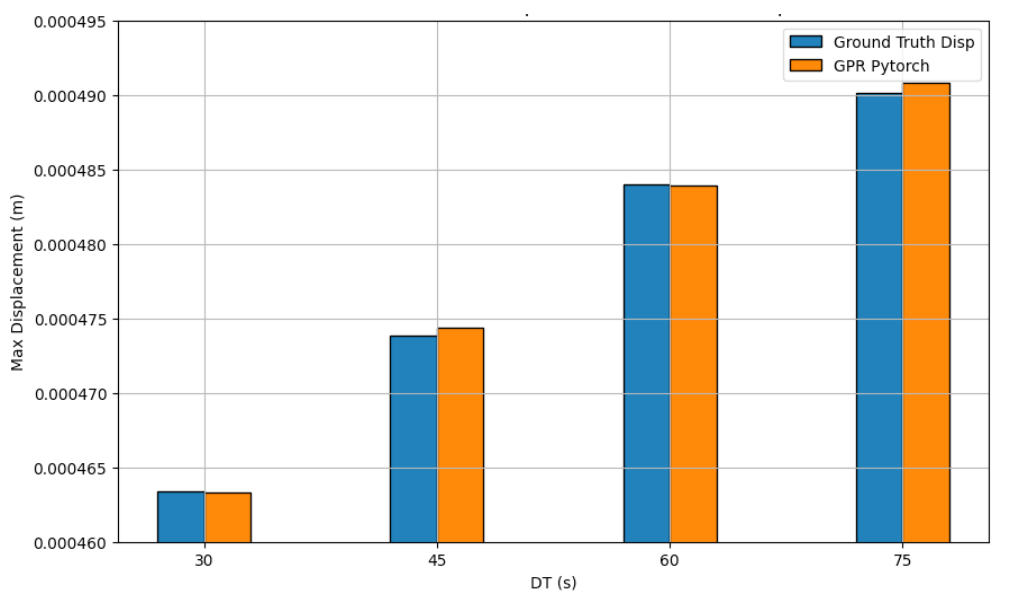}
        \caption{}
        \label{fig:subfig1}
    \end{subfigure}
    \hfill
    \begin{subfigure}{0.49\textwidth}
        \centering
        \includegraphics[width=\textwidth]{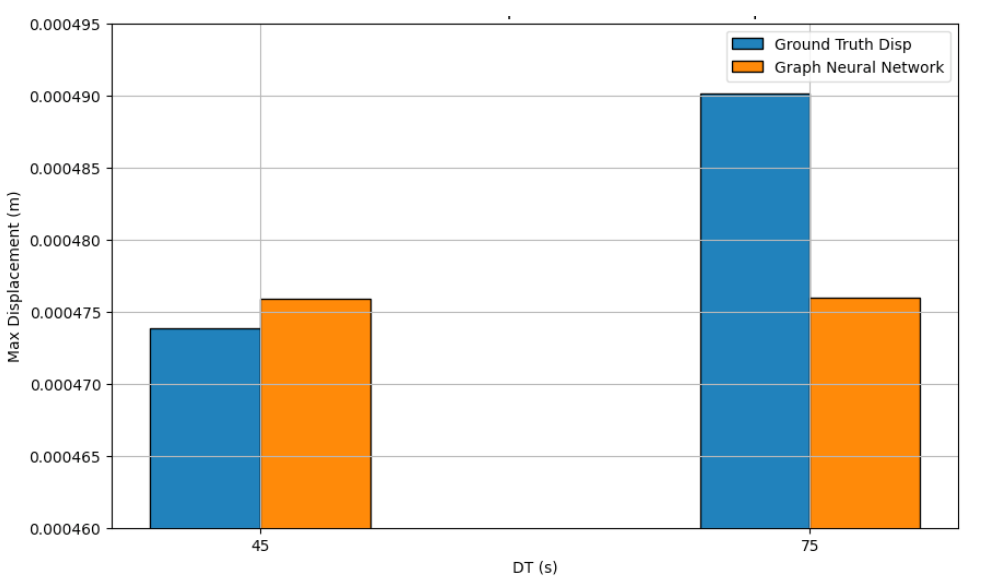}
        \caption{}
        \label{fig:subfig2}
    \end{subfigure}
    \caption{(a) Maximum displacement predictions from the POD-GPR model compared with ground truth from finite element analysis, (b) Predictions from the parameterized GCA model}
    \label{fig:fig3}
\end{figure}

The parameterized GCA was trained within a denoising autoencoder framework, employing early stopping with a patience of 50 epochs and cosine annealing warm restarts \cite{liu2022super} for learning rate adjustment to optimize training and mitigate overfitting. The AdamW optimizer \cite{loshchilov2017fixing} was used for parameter updates. Despite setting the latent space dimension to 12 for a detailed yet compact data representation, the GCA showed tendencies of overfitting, attributed to the limited dataset size of only nine training points. This limited dataset impaired the model's generalization capabilities, particularly noticeable in test performance for dwell times of 45s and 75s. Figure \ref{fig:fig3} contrasts displacement predictions from the POD-GPR and GCA models against finite element simulations. The figure and results underscore the POD-GPR model's superior performance, highlighting its greater accuracy and effectiveness in distortion prediction for additive manufacturing.

\section{Conclusion}
This study highlights the POD-GPR model's exceptional accuracy and computational efficiency in distortion prediction, achieving accuracies within $\pm0.001$ mm and a 1800-fold speed improvement, demonstrating its suitability for engineering applications. While the parameterized GCA model faces challenges in generalizing due to a limited dataset, its versatility in adapting to different geometries indicates significant potential for broader uses. Future work will focus on enhancing the GCA model with an enlarged dataset and exploring advanced non-linear methods such as weak-LaSDI \cite{tran2024weak}.


\begin{ack}
\textbf{I. K.\ Deo, Y.\ Choi, S. A. Khairallah} were supported by Laboratory Directed Research and Development (22-SI-007) Program.
Lawrence Livermore National Laboratory is operated by Lawrence Livermore National Security, LLC, for the U.S. Department of Energy, National Nuclear Security Administration under Contract DE-AC52-07NA27344. IM release number: LLNL-CONF-869106. The authors extend their sincere gratitude to Gabe Guss for their invaluable assistance in conducting experiments and fabricating the 3D-printed cylindrical geometry.
\end{ack}

\bibliographystyle{plainnat}
\bibliography{references}

\end{document}